\documentclass[11pt]{article}

\usepackage[final]{acl}

\usepackage{times}
\usepackage{latexsym}

\usepackage[T1]{fontenc}
\usepackage[utf8]{inputenc}

\usepackage{microtype}
\usepackage{inconsolata}

\usepackage{graphicx}
\usepackage{xcolor}
\usepackage{booktabs}
\usepackage{fancyvrb}
\usepackage{wrapfig}
\usepackage{comment}

\usepackage{url}
\usepackage{hyperref}
\definecolor{acllinkblue}{HTML}{000099}
\hypersetup{
  colorlinks=true,
  citecolor=acllinkblue,
  linkcolor=acllinkblue,
  urlcolor=acllinkblue,
  pdfborder={0 0 0},
}

\usepackage{placeins}

\usepackage{iftex}
\ifPDFTeX
  \usepackage{CJKutf8}
  \newenvironment{JPtext}{\begin{CJK}{UTF8}{min}}{\end{CJK}}
  \newcommand{\jpbold}[1]{{\CJKfamily{goth}\bfseries #1}}
\else
  \usepackage{xeCJK}
  \newenvironment{JPtext}{}{}
  \newcommand{\jpbold}[1]{{\bfseries #1}}
\fi

\newcommand{\promptheading}[1]{%
  \par\smallskip\noindent
  \colorbox{gray!15}{\parbox{\dimexpr\columnwidth-2\fboxsep\relax}{\strut\bfseries #1}}%
  \par\nobreak
}
\newcommand{\taskprompt}[2]{%
  \par\vspace{3pt}\noindent
  \fcolorbox{gray!45}{white}{%
    \parbox{\dimexpr\columnwidth-2\fboxsep-2\fboxrule\relax}{%
      \small\ttfamily\raggedright
      \colorbox{gray!12}{\strut #1}%
      \par\vspace{2pt}
      #2%
    }%
  }%
  \par
}
\DefineVerbatimEnvironment{PromptBox}{Verbatim}{fontsize=\small,frame=single,framesep=4pt,rulecolor=\color{gray!45}}

\title{A Large-Scale Dataset for Cross-Lingual Multi-Jurisdiction Patent Drafting from Patent Cooperation Treaty Applications}

\author{
Trung Phuong Le\textsuperscript{1}, 
Hirofumi Nonaka\textsuperscript{2}, 
Koji Marusaki\textsuperscript{3}, 
Katsuhito Sudoh\textsuperscript{4},
Seiya Kawano\textsuperscript{1,5}
  \\
  \textsuperscript{1}Kyoto Institute of Technology \quad
  \textsuperscript{2}Aichi Institute of Technology \quad \\
  \textsuperscript{3}Nagaoka Institute of Technology \quad
  \textsuperscript{4}Nara Women's University \quad 
  \textsuperscript{5}RIKEN GRP
  \\
  \texttt{le.trung.phuong@vision.is.kit.ac.jp, hnonaka@aitech.ac.jp,} 
  \\
  \texttt{sudoh@ics.nara-wu.ac.jp, kawano@kit.ac.jp}
}

\begin{document}
\maketitle

\begin{abstract}
Securing patent protection across countries requires drafting claims separately for each national patent office, in the relevant language and according to its legal and stylistic conventions.
Patent claims can therefore differ across languages and jurisdictions because of translation, target-office drafting practices, and prosecution-stage changes.
We formulate \emph{cross-lingual patent drafting} as generating plausible candidate claim sets for a specified target language, jurisdiction, and prosecution stage from a source claim set in another language. The task models jurisdiction- and stage-specific transformations beyond content-preserving translation.
We construct a large-scale Japanese--English dataset by linking Japan Patent Office (JPO) and United States Patent and Trademark Office (USPTO) records through 884,179 Patent Cooperation Treaty (PCT)-derived patent families.
Across six released configurations, these families yield 4,067,168 directional claim-set instances; four core configurations form the main benchmark.
% Commercial machine translation and zero- and few-shot prompting capture part of the task, while supervised adaptation provides the strongest overall baseline, showing that task-specific patterns can be learned from linked cross-jurisdictional claim sets.
Commercial machine translation and zero- and few-shot prompting capture part of the task, while supervised adaptation provides the strongest overall baseline, showing that task-specific patterns can be learned from linked cross-jurisdictional claim sets.
\end{abstract}

\section{Introduction}
Protecting an invention internationally requires drafting claims separately for multiple national patent offices, making cross-border claim drafting a central activity of intellectual-property practice.
Patent claims define the legal scope of an invention and are therefore central to patent prosecution and patent strategy~\cite{marco2019patent}.
Drafting patent claims is not merely a matter of translating technical
concepts into legal language. During examination, claims are commonly
narrowed or clarified in response to prior-art and disclosure objections,
and their language and scope may be iteratively refined through
interaction with the examiner~\cite{reilly2018amending}.
Applicants commonly use the Patent Cooperation Treaty (PCT) route to pursue related rights in multiple jurisdictions, but each national office examines the resulting application independently.
Claims derived from the same international application can consequently diverge in language, structure, scope, and number.
We use \emph{prosecution} for the application and examination process at a patent office, and \emph{pre-grant} and \emph{granted} for the two publication stages of an application: the claims as published before examination concludes, and the claims as published if protection is granted. A \emph{claim set} is the complete numbered set of claims in one publication, and a \emph{patent family} groups publications deriving from the same original application.

Patent information processing research has established resources for retrieval, classification, and claim understanding~\cite{lupu2017patent,fujii2007overview,shinmori2003patent}, and recent work has begun to study automatic claim refinement at individual patent offices~\cite{kawano2024claimbrush,uspto2023dataset,epo2025}.
Cross-lingual work, meanwhile, has largely framed related patent documents as machine-translation data and sought meaning-preserving sentence or document alignments~\cite{utiyama2007patent,nagata2024japarapat}.
International prosecution has a further property: a target-jurisdiction claim may express the same invention while also reflecting office-specific drafting conventions and prosecution-stage amendments.
The desired output is thus not always a faithful translation of the source; it can be a legally consequential transformation.

\begin{figure}[t]
\centering
 \includegraphics[width=\columnwidth]{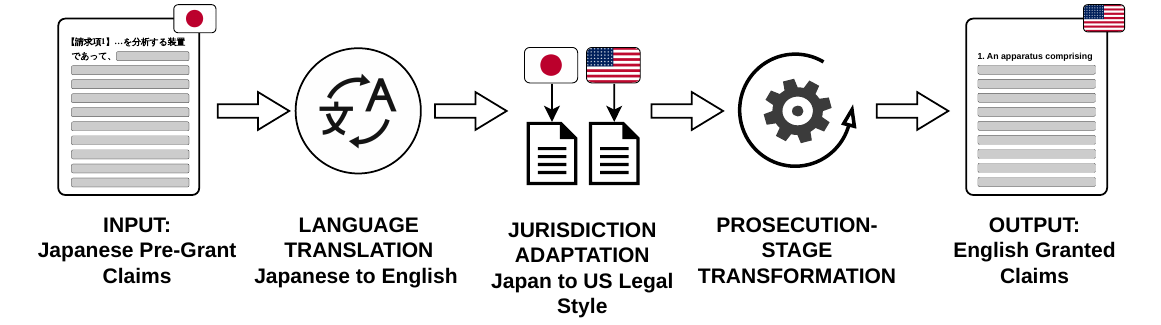}
% \caption{Cross-lingual patent drafting illustrated by the \texttt{ja\_pre$\rightarrow$en\_granted} direction. The task combines language transfer, target-jurisdiction adaptation, and prosecution-stage transformation.}
\caption{Cross-lingual patent drafting and its instantiation in our dataset, illustrated by the \texttt{ja\_pre$\rightarrow$en\_granted} direction. Each task instance links source and target claim sets within the same PCT-derived patent family.}
\label{fig:task_overview}
\end{figure}

% We formulate this problem as \emph{cross-lingual patent drafting}: generating claims in a target language and jurisdiction from claims associated with the same invention in another language.
% We construct a large-scale Japanese--English dataset by linking Japan Patent Office (JPO) and United States Patent and Trademark Office (USPTO) records within patent families and retaining both pre-grant and granted documents.
% The resource contains both \emph{same-stage} variants, in which the source and target are both pre-grant or both granted, and \emph{cross-stage} variants, in which pre-grant claims are used to generate granted claims in the other jurisdiction.
% The release exposes six directional configurations; four core pre-grant-source configurations form the main benchmark, while one granted-to-granted direction is analyzed supplementarily in Appendix~\ref{app:granted_stage}.
% Unlike translation corpora that remove divergent pairs as alignment noise, our resource preserves naturally occurring jurisdictional and prosecution-driven differences as the phenomenon to be modeled.
% Figure~\ref{fig:task_overview} illustrates these transformations for one cross-stage task direction.

Therefore, prior work leaves open how to generate complete patent claim sets when the language, jurisdiction, and prosecution stage change jointly. We address this gap through a new task formulation, a PCT-linked benchmark, and an empirical evaluation. Our contributions are threefold:
\begin{itemize}
    \item We formulate \emph{cross-lingual patent drafting} as complete claim-set generation conditioned on a target language, jurisdiction, and prosecution stage, extending examination-aware claim generation to the cross-lingual and cross-jurisdictional setting.
    \item We construct and release a benchmark from 884,179 PCT-derived Japanese--English patent families, yielding 4,067,168 directional claim-set instances across six configurations with chronological, family-disjoint splits; four pre-grant-source directions form the core benchmark.\footnote{The dataset and evaluation materials are available at \url{https://doi.org/10.6084/m9.figshare.33027056.v1}.}
    \item We establish initial baselines spanning commercial translation, zero- and few-shot prompting, prompt controls, two-stage pipelines, and supervised adaptation. Supervised adaptation obtains the highest BLEU on all four core tasks and the highest SARI on three. In a 25-case pilot on the Japanese granted-stage direction, a human evaluator and an LLM evaluator each prefer its outputs to both evaluated baselines more often than the reverse.
\end{itemize}

\section{Related Work}
\label{sec:related_work}

\paragraph{Cross-lingual patent corpora.}
Cross-lingual patent processing has largely focused on machine translation, supported by parallel corpora derived from patent families and sentence-level alignment~\cite{utiyama2007patent,nagata2024japarapat}.
JaParaPat uses DOCDB family metadata to identify approximately 1.4 million Japanese--US application pairs and extracts sentence alignments for machine translation~\cite{nagata2024japarapat}.
Our resource instead uses complete claim sets as linked task instances, covers both pre-grant and granted publications, and retains differences in claim number, structure, and scope across jurisdictions and prosecution stages as outcomes to be modeled.
Patent NLP has also been organized around evaluation campaigns such as NTCIR, TREC-CHEM, and CLEF-IP, which cover retrieval, classification, and translation~\cite{fujii2007overview,lupu2017patent,trec2010,clef2011}.

\paragraph{Monolingual examination and claim revision.}
A growing line of work studies claim revision and examination within a single jurisdiction, from rule-based analyses of claim structure~\cite{shinmori2003patent} to large-scale revision and examination resources: ClaimBrush (JPO pre-grant-to-granted rewriting)~\cite{kawano2024claimbrush}, Patent-CR (EPO claim revision)~\cite{jiang2025patentcr}, PatentEdits (USPTO edit prediction)~\cite{lee2024patentedits}, PANORAMA (USPTO examination trails)~\cite{lim2025panorama}, HUPD (multi-purpose USPTO applications)~\cite{uspto2023dataset}, and EPD (EPO documents at both pre-grant and granted stages for generation)~\cite{epo2025}.
These resources demonstrate the value of modeling claim revision and examination-stage context, but remain monolingual and single-jurisdiction (Table~\ref{tab:dataset_comparison}).

\paragraph{LLMs in the patent domain.}
Recent surveys document rapid growth of LLM applications to patent retrieval, classification, and generation~\cite{jiang2025survey}, and dedicated studies examine LLMs' ability to draft claims directly~\cite{jiang2024quality}.

% Our contribution is the combination of a new task formulation and an enabling resource: target-jurisdiction claim generation over PCT-linked, claim-set-level correspondences, with both matched-stage and pre-grant-to-granted variants and with naturally occurring divergence preserved.
% Prior corpora may contain related documents from multiple jurisdictions, but they do not jointly operationalize this combination as cross-lingual patent drafting.
% Table~\ref{tab:dataset_comparison} summarizes the positioning.

Prior corpora may contain related documents from multiple jurisdictions, but none jointly operationalizes cross-language generation, target-jurisdiction adaptation, and prosecution-stage transformation at the complete claim-set level.
Table~\ref{tab:dataset_comparison} positions our complete-claim-set benchmark relative to prior patent datasets.

\begin{table*}[t]
\centering
\scriptsize
\setlength{\tabcolsep}{3pt}
\renewcommand{\arraystretch}{1.1}
\resizebox{\textwidth}{!}{%
\begin{tabular}{l|l|r|l|l|l|l|l}
\hline
\textbf{Dataset} & \textbf{Languages} & \textbf{Count} & \textbf{Unit} & \textbf{Task} & \textbf{Stages} & \textbf{Jurisdictions} & \textbf{Granularity} \\
\hline\hline
JaParaPat~\cite{nagata2024japarapat} & JA, EN & $\sim$350M & sentence pairs & translation & pre & JPO, USPTO & sentence \\
NTCIR collections~\cite{fujii2007overview} & JA, EN & --- & documents & retrieval, classif. & mixed & JPO, USPTO & document \\
HUPD~\cite{uspto2023dataset} & EN & 4.52M & applications & multi-purpose & pre & USPTO & document \\
ClaimBrush~\cite{kawano2024claimbrush} & JA & 2.25M & rewriting pairs & refinement & CS & JPO & claim set \\
PatentEdits~\cite{lee2024patentedits} & EN & 105K & draft--grant pairs & edit prediction & CS & USPTO & claim \\
Patent-CR~\cite{jiang2025patentcr} & EN & 22.6K & revision pairs & revision & CS & EPO & claim set \\
PANORAMA~\cite{lim2025panorama} & EN & 8.1K & examination trails & examination & pre, grant & USPTO & document \\
EPD~\cite{epo2025} & EN & 107K & documents & generation & pre, grant & EPO & document \\
\textbf{Ours} & \textbf{JA, EN} & \textbf{884K} & \textbf{families} & \textbf{drafting} & \textbf{pre, grant, CS} & \textbf{JPO, USPTO} & \textbf{claim set} \\
\hline
\end{tabular}
}
\caption{Comparison with existing patent datasets. Counts use each dataset's native unit and are therefore not directly comparable across rows. For ours, 884,179 families yield 4,067,168 directional linked claim-set instances across six released configurations. Pre = pre-grant; grant = granted; CS = cross-stage (pre to grant); a claim set is all claims in one publication. \textbf{Bold} identifies our dataset.}
\label{tab:dataset_comparison}
\end{table*}

\section{Cross-Lingual Patent Drafting Task}
\label{sec:task}

We define \emph{cross-lingual patent drafting} as complete claim-set generation conditioned on a source claim set and a specified target language, jurisdiction, and prosecution stage (Figure~\ref{fig:task_overview}). The output is a plausible candidate claim set for the target setting, rather than a sentence-aligned translation of the source.

We consider two task families. In \emph{same-stage} drafting, the source and target documents are both pre-grant or both granted, so the task combines language transfer with adaptation to the target jurisdiction. In \emph{cross-stage} drafting, pre-grant claims in one jurisdiction are used to generate granted-stage claims in the other, adding prosecution-stage transformation. Unlike monolingual claim revision, these tasks also change the language and jurisdiction.

For cross-stage drafting, the linked granted claim set is one observed prosecution outcome, not a target uniquely determined by the source claims. A source may admit multiple plausible drafts because each office conducts examination independently and patent families can contain one-to-many document correspondences. Models receive only the source claim set; the specification and prosecution history are outside the input. The benchmark therefore measures approximation of patterns found in observed target claim sets, not recovery of a uniquely correct prosecution result.

Operationalizing this task requires complete claim sets linked across languages, jurisdictions, and stages, with their naturally occurring divergence intact. PCT-derived patent families provide these correspondences and motivate the dataset constructed in the following section.

\section{Dataset Construction}
We construct six Japanese--English claim-drafting task directions from paired JPO and USPTO records, four of which form the main benchmark.
The pipeline (1) collects JPO and USPTO records, (2) identifies PCT-linked patent families, (3) selects pre-grant and granted claim sets, (4) applies publication-date and cross-stage visibility constraints, (5) constructs directional task instances, and (6) assigns complete families to chronological train/validation/test splits.\footnote{Appendix~\ref{app:appendix_construction} details the construction procedure.}

\subsection{PCT National-Phase Process and Task Derivation}
\begin{figure}[t]
\centering
\ifPDFTeX
  \includegraphics[width=\columnwidth]{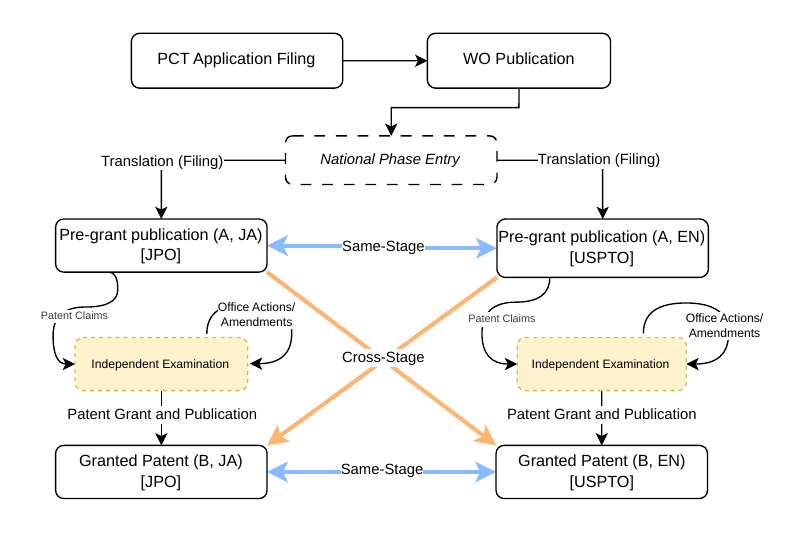}
\else
  \includegraphics[width=\columnwidth]{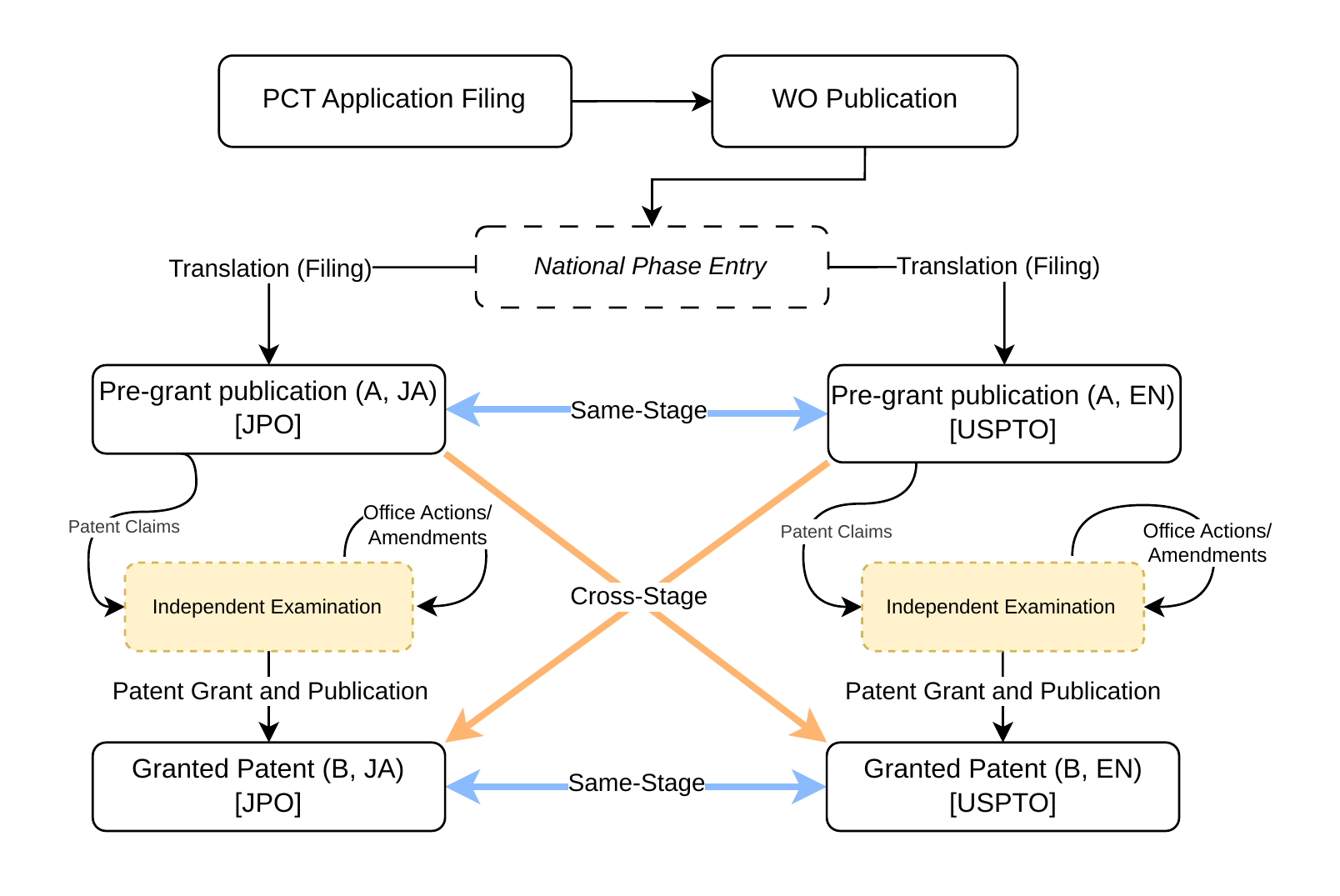}
\fi
\caption{Publication stages and task structure. Same-stage tasks link documents at matched stages across jurisdictions; cross-stage tasks link a pre-grant source to the other jurisdiction's granted claims.}
\label{fig:construction}
\end{figure}

The Patent Cooperation Treaty (PCT) enables applicants to seek patent protection across multiple countries through a single international application.
After filing, applicants enter the ``national phase'' in selected countries, where each patent office examines the application independently.
%We infer stage from office-specific publication kind codes: application-publication codes represent pre-grant records, whereas grant-publication codes represent granted records; the exact A/B suffix conventions differ by office.
We determine each document's stage from its publication kind code, using each office's equivalents of A-type codes (pre-grant application publications) and B-type codes (grant publications).
We use the resulting patent-family groupings and publication stages to form the cross-jurisdictional task instances.
Figure~\ref{fig:construction} illustrates this process and the resulting task structure.

\subsection{Dataset Analysis}
\label{sec:analysis}
The final dataset contains \textbf{4,067,168} directional linked claim-set instances, drawn from \textbf{884,179} distinct patent families, across six released task configurations.
Each instance is a directional linkage between a source-jurisdiction claim set and a target-jurisdiction claim set from the same patent family; it is not a unique application or invention.
Family membership uses the \texttt{family\_id} field from Google Patents Public Data, which groups publications sharing the same priority claims~\cite{googlepatentsdata}.
Construction parameters, including identifier resolution, deduplication, within-family pairing, and length filtering, are given in Appendix~\ref{app:appendix_construction}.
Table~\ref{tab:dataset_stats} reports the final task and split counts.

Several characteristics are noteworthy. 
First, same-stage pre-grant transfer provides 1.26M examples in each direction. 
Second, the two pre-grant-to-granted tasks contribute 817k and 637k examples, respectively, providing substantial supervision for cross-stage drafting. 
Third, the two granted-to-granted tasks are substantially smaller because they require granted claim text in both jurisdictions; we therefore treat them as supplementary tasks. 
Because every instance from a family shares the family's split assignment, the within-family combinatorial expansion cannot leak content between training and evaluation.
Families are assigned to splits in an 80/10/10 ratio; the resulting instance-level proportions are approximately 72/10/18 because the chronological assignment interacts with skewed family sizes and the cross-stage visibility filter.

Beyond these aggregate statistics, an important property of the dataset is that its linked claim sets are not necessarily translation-equivalent.
Unlike claim pairs in conventional patent translation corpora, even pre-grant correspondences across jurisdictions can differ substantially in structure and content because of jurisdiction-specific drafting and prosecution practices.
We therefore preserve these naturally occurring differences rather than filtering them out as noise.

To quantify the extent to which the linked instances diverge from translation equivalence, we compare each reference against a Google-translated version of its source, using chrF~\cite{popovic2015chrf} on the evaluation sample later used across the main and supplementary analyses (Section~\ref{sec:results}).
Overall, 56.0\% of instances score below chrF~50 against the translated source (51.2\% of same-stage and 61.9\% of cross-stage instances).\footnote{Because chrF is computed over characters, values are not comparable across target languages: the median chrF is 57.0--62.3 for English targets and 32.6--33.8 for Japanese targets.}
Moreover, 78.3\% differ from the source in the number of claims, with a median absolute difference of six; among cross-stage instances, the corresponding figures are 90.0\% and seven claims.
Table~\ref{tab:dataset_stats_extended} reports additional statistics for each released task direction.
Granted claim sets contain markedly fewer claims (15.4--19.0 per set) than pre-grant sets (24.8--25.8), further reflecting the structural differences between prosecution stages.

These results indicate that a substantial fraction of the correspondences cannot be treated as straightforward translations, particularly across prosecution stages.
Because the chrF probe relies on machine translation, MT errors alone can lower the score, so the chrF-based figures partly reflect MT error rather than content divergence.

\begin{comment}
Unlike conventional patent translation corpora, even pre-grant correspondences across jurisdictions can differ substantially in structure and content due to jurisdiction-specific practice. Therefore, our dataset preserves these naturally occurring differences rather than filtering them out as noise.

To quantify how far the linked instances diverge from translation equivalence, we compare each reference against a machine translation of its source on the evaluation sample used across the main and supplementary analyses (Section~\ref{sec:exp}).
Overall, 56.0\% of instances score below chrF~50 against the translated source (51.2\% of same-stage and 61.9\% of cross-stage instances), and 78.3\% differ from the source in the number of claims (median absolute difference 6; for cross-stage instances, 90.0\% with a median of 7).
A substantial fraction of correspondences are thus not meaning-preserving translations, particularly across prosecution stages.
Because the probe relies on machine translation, MT errors alone can lower chrF, so these figures may overstate content divergence and should be read as an upper bound.

Table~\ref{tab:dataset_stats_extended} reports extended statistics for each released task direction.
Granted claim sets contain markedly fewer claims (15.4--19.0 per set) than pre-grant sets (24.7--25.8), consistent with the stage gap analyzed in Section~\ref{sec:results}.
\end{comment}

\begin{table*}[t]
\centering
\small
\setlength{\tabcolsep}{6pt}
\renewcommand{\arraystretch}{1.05}
\begin{tabular}{lrrrrr}
\toprule
\textbf{Task direction} & \textbf{Train} & \textbf{Val.} & \textbf{Test} & \textbf{Instances} & \textbf{Families} \\
\midrule
\texttt{ja\_pre$\rightarrow$en\_pre} & 886,840 & 134,693 & 235,715 & 1,257,248 & 834,964 \\
\texttt{en\_pre$\rightarrow$ja\_pre} & 886,840 & 134,693 & 235,715 & 1,257,248 & 834,964 \\
\texttt{ja\_pre$\rightarrow$en\_granted} & 538,392 & 95,492 & 183,467 & 817,351 & 560,858 \\
\texttt{en\_pre$\rightarrow$ja\_granted} & 529,975 & 51,829 & 55,043 & 636,847 & 475,649 \\
\midrule
\texttt{ja\_granted$\rightarrow$en\_granted} & 48,642 & 107 & 488 & 49,237 & 30,581 \\
\texttt{en\_granted$\rightarrow$ja\_granted} & 48,642 & 107 & 488 & 49,237 & 30,581 \\
\midrule
\textbf{Total} & \textbf{2,939,331} & \textbf{416,921} & \textbf{710,916} & \textbf{4,067,168} & \textbf{884,179} \\
\bottomrule
\end{tabular}
\caption{Released task directions and final dataset statistics. The first four directions form the core benchmark; the two granted-to-granted directions below the rule are supplementary, with only \texttt{ja\_granted$\rightarrow$en\_granted} evaluated in Appendix~\ref{app:granted_stage}. Instance counts are directional, and family counts overlap across tasks; the total family count is global and therefore not the sum of rows. Cross-stage tasks retain only instances for which the source was publicly visible before the target grant.}
\label{tab:dataset_stats}
\end{table*}

\begin{table*}[t]
\centering
\scriptsize
\setlength{\tabcolsep}{3pt}
\renewcommand{\arraystretch}{1.05}
\resizebox{\textwidth}{!}{%
\begin{tabular}{l|r|r|r|r|r|r|r|r}
\hline
\textbf{Task direction} & \textbf{Instances} & \textbf{Src vocab.} & \textbf{Tgt vocab.} & \textbf{Src claims} & \textbf{Tgt claims} & \textbf{Mean T/S} & \textbf{Src chars} & \textbf{Tgt chars} \\
\hline\hline
\texttt{ja\_pre$\rightarrow$en\_pre} & 1.26M & 1.15M & 2.30M & 25.82 & 25.77 & 1.20 & 2490 & 5321 \\
\texttt{en\_pre$\rightarrow$ja\_pre} & 1.26M & 2.30M & 1.15M & 25.77 & 25.82 & 1.41 & 5321 & 2490 \\
\texttt{ja\_pre$\rightarrow$en\_granted} & 817.4K & 886.9K & 1.52M & 25.80 & 17.01 & 0.93 & 2546 & 5248 \\
\texttt{en\_pre$\rightarrow$ja\_granted} & 636.8K & 1.47M & 665.9K & 24.79 & 15.39 & 0.88 & 5416 & 1985 \\
\texttt{ja\_granted$\rightarrow$en\_granted} & 49.2K & 108.3K & 211.7K & 15.62 & 18.97 & 1.97 & 1713 & 4560 \\
\texttt{en\_granted$\rightarrow$ja\_granted} & 49.2K & 211.7K & 108.3K & 18.97 & 15.62 & 1.35 & 4560 & 1713 \\
\hline
\end{tabular}
}
\caption{Extended statistics for the six released task directions. Vocabulary size uses MeCab (IPA dictionary) for Japanese and whitespace-after-punctuation splitting for English. Claims are means per claim set; Mean T/S is the mean per-instance target/source claim-count ratio. Character columns are median NFC-normalized Unicode code-point counts.}
\label{tab:dataset_stats_extended}
\end{table*}

\subsection{Example Claim-Level Comparison}
\label{sec:qual_examples}

% To illustrate the difference between translation-style correspondences and drafting-stage changes, Table~\ref{tab:claim_gap} shows Claim~1 of four national-phase documents derived from PCT publication \texttt{WO2001/095404}: the Japanese and English pre-grant publications and the corresponding granted publications.
To illustrate the difference between translation-style correspondences and drafting-stage changes, Table~\ref{tab:claim_gap} shows Claim~1 of four national-phase documents derived from PCT publication \texttt{WO2001/095404}: the JPO and USPTO pre-grant publications and their corresponding grants. The JPO claims are shown in English translation (original Japanese in Appendix~\ref{app:claim_gap_japanese}).

% On both language sides, the granted claims add limitations absent from the pre-grant versions: the Japanese granted claim introduces a lattice frame with through-holes and heater wires, and the English granted claim adds recessed flow-distribution plates with communication passages.
On both jurisdictional sides, the granted claims add limitations absent from the pre-grant versions: the JPO granted claim specifies anodic bonding by electrode and glass films together with an embedded heater, and the USPTO granted claim adds recessed flow-distribution plates with communication passages.
% The two pre-grant claims also emphasize different aspects of the invention (airtight frame bonding vs.\ planar cell arrangement), reflecting claim reorganization during separate national-phase entry, an instance of the jurisdiction-specific divergence discussed in Section~\ref{sec:analysis}.
The two pre-grant claims also emphasize different aspects of the invention (airtight frame bonding vs.\ planar cell arrangement), an instance of the jurisdiction-specific divergence discussed in Section~\ref{sec:analysis}.
This illustrates the intended distinction in our dataset: cross-stage drafting requires modeling prosecution-stage changes beyond cross-lingual transfer.

\begin{table*}[t]
\centering
\scriptsize
\setlength{\tabcolsep}{5pt}
\renewcommand{\arraystretch}{1.1}
\begin{tabular}{l|p{0.42\linewidth}|p{0.42\linewidth}}
\hline
 % & \textbf{Japanese} & \textbf{English} \\
 & \textbf{JPO (English translation)} & \textbf{USPTO (original English)} \\
\hline\hline
\textbf{Pre-grant} &
\texttt{JP-2004501492-A} (2004-01-15), Claim~1
\newline
% \begin{JPtext}電解質層(2)と、前記電解質層を挟む一対のガス拡散電極層(3,4)と、前記各ガス拡散電極層に接する燃料ガス及び酸化剤ガスの通路(10,11)を画定するための一対の配流板(5)とを有する少なくとも1つのセルから構成された燃料電池であって、前記電解質層(2)が、フレーム(21)及び該フレーム内に保持された電解質(22)を有し、前記両配流板と前記フレームとが、内部が気密になるようにその外周全周に亘り接合または接着されていることを特徴とする燃料電池。\end{JPtext}
A fuel cell comprising at least one cell having an electrolyte layer (2), a pair of gas-diffusion electrode layers (3, 4) sandwiching the electrolyte layer, and a pair of flow-distribution plates (5) defining passages (10, 11) for fuel gas and oxidant gas that contact the respective gas-diffusion electrode layers, wherein the electrolyte layer (2) has a frame (21) and an electrolyte (22) held within the frame, and the two flow-distribution plates and the frame are joined or bonded around their entire outer peripheries so as to make the interior airtight.
&
\texttt{US-2004091765-A1} (2004-05-13), Claim~1 \newline
A fuel cell assembly comprising a plurality of cells each including an electrolyte layer, a pair of diffusion electrode layers interposing said electrolyte layer between them, and a pair of flow distribution plates for defining passages for fuel and oxidant fluids that contact said diffusion electrode layers, wherein: said fuel cells are arranged on a common plane.
\\
\hline
\textbf{Granted} &
\texttt{JP-4989802-B2} (2012-08-01), Claim~1 \newline
% \begin{JPtext}電解質層(2)と、前記電解質層を挟む一対のガス拡散電極層(3,4)と、前記各ガス拡散電極層に接する燃料ガス及び酸化剤ガスの通路(10,11)を画定するための一対の配流板(5)とを有する少なくとも1つのセルから構成された燃料電池であって、前記電解質層(2)が、フレーム(21)及び該フレーム内に保持された電解質(22)を有し、前記両配流板と前記フレームとが、内部が気密になるようにその外周全周に亘り、\jpbold{前記両配流板の外周部に成膜された電極膜と前記フレームの外周部に成膜されたガラス膜とにより陽極接合され、前記電極膜の下層に陽極接合のためのヒータが埋設されている}ことを特徴とする燃料電池。\end{JPtext}
A fuel cell comprising at least one cell having an electrolyte layer (2), a pair of gas-diffusion electrode layers (3, 4) sandwiching the electrolyte layer, and a pair of flow-distribution plates (5) defining passages (10, 11) for fuel gas and oxidant gas that contact the respective gas-diffusion electrode layers, wherein the electrolyte layer (2) has a frame (21) and an electrolyte (22) held within the frame, and the two flow-distribution plates and the frame are, around their entire outer peripheries, \textbf{anodically bonded by an electrode film formed on the outer peripheral portions of the two flow-distribution plates and a glass film formed on the outer peripheral portion of the frame, with a heater for anodic bonding embedded beneath the electrode film}, so as to make the interior airtight.
&
\texttt{US-6991868-B2} (2006-01-31), Claim~1 \newline
A fuel cell assembly comprising a plurality of cells each including an electrolyte layer, a pair of diffusion electrode layers interposing said electrolyte layer between them, and a pair of flow distribution plates for defining passages for fuel and oxidant fluids that contact said diffusion electrode layers, wherein: said fuel cells are arranged on a common plane, \textbf{wherein each of said flow distribution plates comprises a plurality of recesses formed on a side thereof facing said electrolyte layer so as to form individual cells, said recesses being communicated to one another according to a prescribed pattern by said communication passages.}
\\
\hline
\end{tabular}
% \caption{Claim-level comparison for one invention across both jurisdictions and prosecution stages. All four documents derive from PCT publication \texttt{WO2001/095404}, prosecuted independently at the JPO and USPTO. Bolded phrases highlight limitations added during each office's examination.}
\caption{English-language claim comparison for one invention across jurisdictions and prosecution stages. The JPO claims are translated into English for presentation; the original Japanese appears in Appendix~\ref{app:claim_gap_japanese}. All four documents derive from PCT publication \texttt{WO2001/095404} and were prosecuted independently at the JPO and USPTO. Bold text highlights limitations added during examination.}

\label{tab:claim_gap}
\end{table*}

\section{Experimental Setup}
\label{sec:exp}
%To establish initial reference points and test whether existing approaches can produce the required jurisdictional and prosecution-stage transformations, we benchmark translation, prompting, pipeline, and adaptation approaches on the four core task directions---\texttt{ja\_pre$\rightarrow$en\_pre}, \texttt{en\_pre$\rightarrow$ja\_pre}, \texttt{ja\_pre$\rightarrow$en\_granted}, and \texttt{en\_pre$\rightarrow$ja\_granted}---in the source-only setting of Section~\ref{sec:task}.
%Appendix~\ref{app:granted_stage} reports \texttt{ja\_granted$\rightarrow$en\_granted} as a supplementary analysis.

To establish initial reference points and test whether existing approaches can produce the required jurisdictional and prosecution-stage transformations, we benchmark translation, prompting, pipeline, and adaptation approaches on the four core task directions (\texttt{ja\_pre$\rightarrow$en\_pre}, \texttt{en\_pre$\rightarrow$ja\_pre}, \texttt{ja\_pre$\rightarrow$en\_granted}, and \texttt{en\_pre$\rightarrow$ja\_granted}) in the source-only setting of Section~\ref{sec:task}.\footnote{Appendix~\ref{app:granted_stage} reports \texttt{ja\_granted$\rightarrow$en\_granted} as a supplementary analysis.} 

\subsection{Evaluation Set}

From the family-disjoint held-out test partition, we sample 1,000 instances from each of the four core task directions, yielding 4,000 evaluation instances in total. All systems are evaluated on the identical instances. The training partition is used to fit the adapter and the validation partition is used to monitor training.

\subsection{Systems}
We compare the following systems; the full prompt texts are provided in Appendix~\ref{app:prompts}.

\begin{itemize}
\item \textbf{Google Translate v3}: Google Cloud Translation API (Advanced) v3\footnote{\url{https://docs.cloud.google.com/translate/docs/api-overview\#advanced_ed}} provides a strong commercial machine-translation baseline.
\item \textbf{Qwen3.6-27B zero-/few-shot}~\cite{qwen2026qwen36}\footnote{\url{https://huggingface.co/Qwen/Qwen3.6-27B}}: the zero-shot condition uses a task-specific patent-drafting instruction specifying the source and target languages and prosecution stages; the few-shot condition adds one manually selected example from the same task direction, held fixed across all evaluation instances in that direction (Appendix~\ref{app:prompts}). As a paired prompt-control diagnostic, we additionally rerun the zero-shot condition on the same 1,000 \texttt{en\_pre$\rightarrow$ja\_granted} examples with the \texttt{en\_pre$\rightarrow$ja\_pre} same-stage instruction, keeping source texts, preprocessing, and Japanese granted references identical so that only the requested target stage changes.
\item \textbf{Qwen3.6-27B simple-MT}: an ablation that replaces the drafting instruction with a plain translation request, isolating the effect of task framing.
\item \textbf{OpenAI GPT-5.4 zero-/few-shot (API)}: the zero-shot condition uses the same patent-drafting instruction and prompt template as Qwen3.6-27B zero-shot; the few-shot condition adds the same task-specific demonstration used for Qwen. These conditions test whether a stronger general-purpose prompted model can perform the required jurisdictional and prosecution-stage transformations.
\item \textbf{GPT-5.4 two-stage pipelines}: To test whether cross-lingual drafting decomposes into monolingual drafting plus translation, we evaluate two pipelines on the cross-stage tasks. \emph{Draft-then-Translate} first performs monolingual pre-grant-to-granted drafting in the source language and then translates the result; \emph{Translate-then-Draft} first translates the pre-grant claims and then performs monolingual drafting in the target language. The drafting prompts follow the monolingual claim-rewriting framing of \citet{kawano2024claimbrush}, and the translation stage reuses the simple-MT prompt.
\item \textbf{Qwen3.6-27B QLoRA}~\cite{dettmers2023qlora}: a supervised-adaptation baseline that tests whether task-specific transformations can be learned from cross-jurisdictional supervision. We train one shared multitask adapter, rather than separate task-specific adapters, on 100,000 instances drawn from the family-disjoint training partition: 20,000 from each of the four core tasks and 20,000 from the supplementary \texttt{ja\_granted$\rightarrow$en\_granted} task. Full training and inference settings are provided in Appendix~\ref{app:prompts}.
\end{itemize}

\subsection{Automatic and Human Evaluation}
\paragraph{Automatic evaluation.}
We report three automatic metrics:

\begin{itemize}
\item \textbf{BLEU}~\cite{papineni2002bleu}: exact token-sequence overlap with the reference.
\item \textbf{chrF}~\cite{popovic2015chrf}: a character-level view that is useful for Japanese generation.
\item \textbf{SARI}~\cite{xu2016sari}: rewards appropriate additions, deletions, and retained content relative to the source and reference.
\end{itemize}

Because SARI assumes monolingual editing, we replace each source claim set with its Google-translated version in the target language before scoring, placing the source, prediction, and reference in the same language; Google Translate's own output therefore serves as the SARI no-edit baseline.
We evaluate complete claim sets from the final dataset (Section~\ref{sec:analysis}) and apply no additional length filtering to the benchmark.

For the cross-stage tasks, we additionally report mean absolute differences in claim count from the reference ($|\Delta_r|$, lower is better) and from the source ($|\Delta_s|$).

\paragraph{Pilot pairwise evaluation.}
To complement the automatic metrics, we conduct a blind pairwise evaluation on 25 sampled \texttt{en\_pre$\rightarrow$ja\_granted} cases: outputs from Qwen simple-MT, GPT-5.4 zero-shot, and Qwen QLoRA are compared in all three pairings (75 judgments), with system identities hidden, candidate order randomized, and ties permitted. Two evaluators judge each pair independently, selecting the candidate that could be used in practice as a Japanese granted-stage claim set, as-is or with less revision. 
The human evaluator has professional expertise in JPO practice; because it was difficult to secure an evaluator versed in both JPO and USPTO practice, the pilot is restricted to the direction in which JPO expertise directly applies. 
The LLM evaluator (Anthropic Fable 5\footnote{\texttt{claude-fable-5}, accessed July 2026; \url{https://www.anthropic.com/claude}}, default decoding parameters) receives the same instructions and has no access to the human judgments.
Judges saw the English source, an auxiliary Japanese pre-grant claim set, and two anonymized outputs; the observed Japanese granted reference was not shown. 
Appendix~\ref{app:human_eval} provides the complete instructions, evaluation criteria, eligibility filters, sampling procedure, and randomization details.

\section{Experimental Results}
\label{sec:results}
%Table~\ref{tab:results_all} reports results on the evaluation set; values are rounded to two decimals.
Table~\ref{tab:results_all} reports results on the evaluation set.

\begin{table*}[t]
\centering
\footnotesize
\setlength{\tabcolsep}{2.0pt}
\renewcommand{\arraystretch}{1.04}
\resizebox{\textwidth}{!}{%
\begin{tabular}{@{}lrrrrrrrrrrrr@{}}
\toprule
\textbf{System}
& \multicolumn{3}{c}{\textbf{JA-P$\rightarrow$EN-P}}
& \multicolumn{3}{c}{\textbf{EN-P$\rightarrow$JA-P}}
& \multicolumn{3}{c}{\textbf{JA-P$\rightarrow$EN-G}}
& \multicolumn{3}{c}{\textbf{EN-P$\rightarrow$JA-G}} \\
\cmidrule(lr){2-4}\cmidrule(lr){5-7}\cmidrule(lr){8-10}\cmidrule(l){11-13}
& BLEU$\uparrow$ & chrF$\uparrow$ & SARI$\uparrow$
& BLEU$\uparrow$ & chrF$\uparrow$ & SARI$\uparrow$
& BLEU$\uparrow$ & chrF$\uparrow$ & SARI$\uparrow$
& BLEU$\uparrow$ & chrF$\uparrow$ & SARI$\uparrow$ \\
\midrule
Google Translate & 24.81 & 58.97 & 48.00 & 18.60 & 27.35 & 41.48 & 21.72 & 54.77 & 46.81 & 24.39 & 32.48 & 39.48 \\
\midrule
Qwen zero-shot & 31.08 & 61.71 & 49.30 & 18.30 & 27.29 & 59.66 & 28.01 & 57.61 & 48.65 & 26.35 & 32.45 & 57.48 \\
Qwen few-shot & 31.69 & 61.84 & 49.49 & 17.87 & 27.13 & 61.19 & 27.51 & 57.37 & 48.53 & 29.58 & 34.68 & 58.82 \\
Qwen simple-MT & 30.87 & 61.55 & 49.01 & 18.43 & 27.49 & 59.86 & 26.94 & 57.19 & 47.72 & 26.35 & 32.93 & 56.98 \\
\midrule
GPT-5.4 zero-shot & 31.24 & 62.42 & 50.72 & 17.58 & 26.93 & 58.03 & 28.18 & \textbf{58.57} & \textbf{50.40} & 24.74 & 30.88 & 58.88 \\
GPT-5.4 few-shot & 31.93 & \textbf{62.43} & 50.96 & 18.90 & 28.65 & 57.91 & 28.24 & 58.12 & 50.19 & 28.35 & 33.78 & 58.80 \\
\shortstack[l]{GPT-5.4 Draft-then-Translate} & -- & -- & -- & -- & -- & -- & 28.75 & 56.23 & 47.81 & 29.66 & 34.73 & 57.86 \\
\shortstack[l]{GPT-5.4 Translate-then-Draft} & -- & -- & -- & -- & -- & -- & 27.55 & 57.36 & 48.80 & 26.01 & 32.02 & 57.78 \\
\midrule
Qwen QLoRA & \textbf{35.77} & 60.26 & \textbf{52.01} & \textbf{23.73} & \textbf{31.74} & \textbf{62.76} & \textbf{33.73} & 54.86 & 50.29 & \textbf{32.53} & \textbf{36.28} & \textbf{61.62} \\
\bottomrule
\end{tabular}%
}

\vspace{0.6em}

\begin{tabular*}{\textwidth}{@{\extracolsep{\fill}}lrrrr@{}}
\toprule
\textbf{System}
& \multicolumn{2}{c}{\textbf{JA-P$\rightarrow$EN-G}}
& \multicolumn{2}{c}{\textbf{EN-P$\rightarrow$JA-G}} \\
\cmidrule(lr){2-3}\cmidrule(l){4-5}
& $|\Delta_r|\downarrow$ & $|\Delta_s|$ & $|\Delta_r|\downarrow$ & $|\Delta_s|$ \\
\midrule
Reference & 0.00 & 16.42 & 0.00 & 17.51 \\
\midrule
Google Translate & 14.57 & 4.10 & 14.95 & 4.02 \\
\midrule
Qwen zero-shot & 12.39 & 4.37 & 10.84 & 10.38 \\
Qwen few-shot & 12.33 & 4.44 & 9.43 & 11.12 \\
Qwen simple-MT & 12.21 & 4.58 & 16.55 & 1.85 \\
\midrule
GPT-5.4 zero-shot & 13.95 & 5.38 & 11.88 & 14.52 \\
GPT-5.4 few-shot & 13.31 & 4.65 & \textbf{9.02} & 12.17 \\
\shortstack[l]{GPT-5.4 Draft-then-Translate} & \textbf{9.44} & 8.67 & 13.92 & 6.06 \\
\shortstack[l]{GPT-5.4 Translate-then-Draft} & 12.44 & 5.63 & 11.70 & 13.68 \\
\midrule
Qwen QLoRA & 10.35 & 10.72 & 9.94 & 12.97 \\

\bottomrule
\end{tabular*}
\caption{Unified benchmark results. The upper panel reports overlap and edit metrics for the four core tasks; the lower panel reports mean absolute claim-count differences from the reference ($|\Delta_r|$, lower is better) and source ($|\Delta_s|$) on the cross-stage tasks. P and G denote pre-grant and granted stages; dashes indicate pipelines not evaluated on same-stage tasks. The Reference row records the references' inherent $|\Delta_s|$. Bold marks the best system value.}
\label{tab:results_all}
\end{table*}

\paragraph{Prompting helps selectively.}
Google Translate obtains the lowest BLEU in three of the four core tasks, including both cross-stage directions, trailing the best system by 8.1--12.0 BLEU and supporting the premise that the observed target claims require more than translation alone.
However, Qwen's task-specific drafting prompt differs from its simple-MT prompt by at most 1.07 BLEU and 0.48 chrF across the four tasks, and changing only the requested target stage in the paired prompt control shifts scores by only 0.05 BLEU and 0.52 chrF (Table~\ref{tab:diagnostics}a), suggesting that prompting does not reliably induce the required jurisdictional and prosecution-stage transformations.

GPT-5.4 Draft-then-Translate is a notable exception, improving BLEU by 4.92 points over GPT-5.4 zero-shot on \texttt{en\_pre$\rightarrow$ja\_granted}.
One possible explanation is that rewriting before translation separates prosecution-stage revision from cross-lingual transfer, although the present experiment does not establish this mechanism.
However, its generated claim counts remain closer to the source than to the reference ($|\Delta_s|=6.06$ versus $|\Delta_r|=13.92$; Table~\ref{tab:results_all}, lower panel), suggesting that the BLEU gain may reflect improved wording rather than complete structural transformation.

\paragraph{Adaptation improves BLEU across all core tasks.}
QLoRA obtains the highest BLEU in all four core tasks, exceeding the best non-adapted system by 2.87--4.98 BLEU, and leads SARI in three, trailing GPT-5.4 zero-shot by only 0.11 on \texttt{ja\_pre$\rightarrow$en\_granted}.
Relative to Qwen zero-shot, it also reduces $|\Delta_r|$ on both cross-stage tasks while increasing $|\Delta_s|$ toward the references' inherent shifts of 16.42 and 17.51.
These results suggest that supervised cross-jurisdictional examples provide task-specific information that is not reliably elicited through prompting.
On the two core English-target tasks, QLoRA's outputs are 11.6\% and 21.6\% shorter and trade character recall for precision (chrP $+$3.72, chrR $-$3.54 on \texttt{ja\_pre$\rightarrow$en\_pre}), explaining why BLEU improves while chrF falls; Appendix~\ref{app:diagnostics} reports the precision/recall decomposition.

\paragraph{Claim-count agreement differs from textual overlap.}
System rankings under the claim-count metrics in the lower panel of Table~\ref{tab:results_all} differ from the BLEU and SARI rankings: GPT-5.4 few-shot attains the best $|\Delta_r|$ on \texttt{en\_pre$\rightarrow$ja\_granted} despite mid-range BLEU, while simple-MT barely alters the source structure yet leaves the largest reference gap. 
% A stronger textual overlap therefore does not imply a successful restructuring of the claim set.
Stronger textual overlap therefore does not imply closer claim-count agreement.

Cross-stage behavior is also direction-dependent: Draft-then-Translate attains the best $|\Delta_r|$ on \texttt{ja\_pre$\rightarrow$en\_granted} but is among the worst on the other cross-stage direction. However, current experiments do not isolate the effects of target language, jurisdiction, and independently developed prosecution history, so the cause remains open.
Even the strongest results leave mean claim-count gaps of 9.0--9.4 claims, indicating substantial scope to improve structural control.
% Appendix~\ref{app:qualitative} contrasts Google Translate and QLoRA on two aligned cross-stage cases: a selected consolidation success and an over-consolidation failure, illustrating that supervised adaptation can produce reference-like restructuring but can also discard branches retained by the observed grant.
Appendix~\ref{app:qualitative} contrasts Google Translate and QLoRA on three aligned cross-stage cases: a consolidation success, a source-supported claim-narrowing success, and an over-consolidation failure. Together, they illustrate that supervised adaptation can produce reference-like restructuring, but can also discard branches retained by the observed grant.

\paragraph{Pilot pairwise preferences.}
Table~\ref{tab:human_eval} reports both evaluators, who produce the same preference ordering (QLoRA $>$ GPT-5.4 zero-shot $>$ simple-MT), with weaker margins for the LLM evaluator. Human--LLM agreement, computed over the three labels (first candidate, second candidate, and tie), is 82.7\% (Cohen's $\kappa=0.68$~\cite{cohen1960coefficient}), rising to 88.4\% ($\kappa=0.77$) when tie-involving comparisons are excluded. Disagreements concentrate in the QLoRA--GPT-5.4 comparison, where the LLM evaluator chose GPT-5.4 or a tie more often than the human evaluator. 
On these 25 cases, both evaluators preferred QLoRA outputs over each evaluated baseline more often than the reverse.
Item-level judgments are provided in the released repository.

\begin{table}[t]
\centering
\footnotesize
\setlength{\tabcolsep}{2pt}
\resizebox{\columnwidth}{!}{%
\begin{tabular}{@{}llccccc@{}} % rrrrr, ccccc
\toprule
\textbf{Comparison} & \textbf{Ev.} & \textbf{QLoRA} & \textbf{GPT} & \textbf{MT} & \textbf{Tie} & \textbf{Agr.} \\
\midrule
QLoRA--MT  & H & 20 (80\%) & --        & 5 (20\%) & 0        & 96\% \\
           & L & 19 (76\%) & --        & 6 (24\%) & 0        &      \\
\addlinespace[2pt]
QLoRA--GPT & H & 19 (76\%) & 4 (16\%)  & --       & 2 (8\%)  & 64\% \\
           & L & 13 (52\%) & 8 (32\%)  & --       & 4 (16\%) &      \\
\addlinespace[2pt]
GPT--MT    & H & --        & 19 (76\%) & 6 (24\%) & 0        & 88\% \\
           & L & --        & 16 (64\%) & 8 (32\%) & 1 (4\%)  &      \\
\bottomrule
\end{tabular}%
}
\caption{Blind pairwise preferences. H and L denote the human and LLM evaluators; GPT denotes GPT-5.4 zero-shot and MT denotes Qwen simple-MT. Percentages use the 25 cases in each comparison, and Agr.\ is per-comparison human--LLM agreement over the three labels (first candidate, second candidate, tie).}
\label{tab:human_eval}
\end{table}

\section{Conclusion}
We introduced cross-lingual patent drafting as generating plausible candidate claim sets for a specified target language, jurisdiction, and prosecution stage, supported by a large-scale Japanese--English resource spanning pre-grant and granted claims.
Our diagnostic experiments show that translation and prompting capture part of the task, while supervised adaptation achieves stronger overall performance and claim counts closer to the reference.
These results indicate that linked cross-jurisdictional claim sets contain learnable task-specific patterns and provide a foundation for future systems that support professional drafting and review.
Future work includes: (1) extending the dataset to cover additional jurisdictions such as the EPO and other major patent offices, (2) developing specialized models for cross-lingual patent drafting that can learn jurisdiction-specific adaptation patterns, and (3) incorporating examination history information to guide the drafting process across languages.

\section*{Limitations}

The source claim set alone does not explain the amendments reflected in a target grant.
Specifications and prosecution records provide additional evidence needed to assess the
support for generated changes.

Each evaluation instance has one observed target claim set, although several drafts may be plausible, so reference-based scores capture only part of drafting quality. Expert human assessment remains limited in scope, and establishing professional usefulness more broadly requires further evaluation.

The dataset covers Japanese--English PCT-linked applications at the JPO and USPTO. Language and jurisdiction vary together in this setting, so their effects cannot be isolated, and generalization to other filing routes and patent offices remains untested.

The benchmark uses public patent documents that may occur in model pretraining data.
Family-disjoint task splits do not establish separation from pretraining corpora.

These constraints point to several directions for future work: evidence-grounded and retrieval-augmented drafting that conditions on specifications and prosecution histories and verifies whether generated modifications are technically supported; claim-level alignment enabling finer-grained supervision and controllable generation of claim count, dependency structure, and supported limitations; and temporally held-out or access-controlled evaluation that tests generalization to genuinely new inventions, combining automatic analysis with expert qualitative assessment. This benchmark is a first step toward evidence-grounded, structurally controllable, cross-jurisdictionally generalizable, and professionally evaluated cross-lingual patent drafting.
\section*{Ethics Statement}
The dataset is derived exclusively from patent documents published by the JPO and USPTO.
These are official public records, and the extracted content contains no personal information beyond bibliographic data already disclosed by the patent offices.
Automatically generated patent claims are not legal advice: systems trained on this resource are intended to assist, not replace, qualified patent professionals, and outputs require expert review before any legal use.
The USPTO-side text derives from "Google Patents Public Data" by IFI CLAIMS Patent Services and Google~\cite{googlepatentsdata}, licensed under CC~BY~4.0, which permits redistribution of derived data with attribution.
The JPO-side text is a subset of the dataset constructed for the ClaimBrush project~\cite{kawano2024claimbrush}, which derives from the Japan Patent Office's publicly provided bulk data, and was provided to us by the ClaimBrush authors with permission for redistribution.
The complete dataset is available at the repository below, subject to a non-commercial-use restriction, together with documentation of its construction procedure.\footnote{\url{https://doi.org/10.6084/m9.figshare.33027056.v1}}

% Review version: acknowledgments are omitted for anonymity.
% For the final version, uncomment this block immediately before the references:
\section*{Acknowledgments}
This work was supported by NEDO wakasapo project (25101271-0), Okawa Foundation Research Grant and JST ACT-X (JPMJAX22A4).

\bibliography{custom}

\appendix
\section{Supplementary Granted-Stage Analysis}
\label{app:granted_stage}

The main benchmark focuses on the four larger directions with pre-grant sources. This appendix evaluates \texttt{ja\_granted$\rightarrow$en\_granted} using all 488 available test instances from 424 families.

Table~\ref{tab:granted_stage_supp} compares this task with the matched-direction \texttt{ja\_pre$\rightarrow$en\_pre} task. The granted-stage task produces higher BLEU and SARI but lower chrF for every scored generation system. This pattern may reflect more stabilized structure and terminology after examination and amendment, together with the shorter length of granted claims.

\begin{table*}[t]
\centering
\small
\setlength{\tabcolsep}{5pt}
\renewcommand{\arraystretch}{1.05}
\begin{tabular}{lrrrrrr}
\toprule
& \multicolumn{3}{c}{\textbf{JA-P$\rightarrow$EN-P}} & \multicolumn{3}{c}{\textbf{JA-G$\rightarrow$EN-G}} \\
\cmidrule(lr){2-4}\cmidrule(l){5-7}
\textbf{System} & BLEU$\uparrow$ & chrF$\uparrow$ & SARI$\uparrow$ & BLEU$\uparrow$ & chrF$\uparrow$ & SARI$\uparrow$ \\
\midrule
Google Translate & 24.81 & 58.97 & 48.00 & 30.52 & 56.16 & 51.44 \\
Qwen zero-shot & 31.08 & 61.71 & 49.30 & 35.59 & 58.63 & 54.93 \\
Qwen few-shot & 31.69 & 61.84 & 49.49 & 35.48 & 58.54 & 55.00 \\
Qwen simple-MT & 30.87 & 61.55 & 49.01 & 35.40 & 58.57 & 54.25 \\
GPT-5.4 zero-shot & 31.24 & 62.42 & 50.72 & 36.04 & 59.09 & 56.23 \\
GPT-5.4 few-shot & 31.93 & \textbf{62.43} & 50.96 & \textbf{36.84} & \textbf{59.48} & 56.51 \\
\midrule
Qwen QLoRA & \textbf{35.77} & 60.26 & \textbf{52.01} & 36.09 & 58.30 & \textbf{56.85} \\
\bottomrule
\end{tabular}
\caption{Supplementary comparison of Japanese-to-English pre-grant and granted-stage transfer. P and G denote pre-grant and granted stages; SARI scoring follows Section~\ref{sec:exp}. Bold marks the best system within each task and metric.}
\label{tab:granted_stage_supp}
\end{table*}
\section{Additional Diagnostics}
\label{app:diagnostics}

Table~\ref{tab:diagnostics} records the paired target-stage prompt control and the character precision/recall decomposition used in Section~\ref{sec:results}.
Panel~(a) is a separately generated paired rerun; its cross-stage row is the internal baseline for the prompt comparison, not the independently saved Qwen zero-shot benchmark output in Table~\ref{tab:results_all}.

\begin{table*}[t]
\centering
\small
\begin{minipage}[t]{0.39\textwidth}
\centering
\textbf{(a) Paired prompt control: EN-P$\rightarrow$JA-G}\\[3pt]
\begin{tabular}{lrr}
\toprule
\textbf{Instruction} & BLEU$\uparrow$ & chrF$\uparrow$ \\
\midrule
Cross-stage target & 26.26 & 32.39 \\
Same-stage target & 26.31 & 32.91 \\
\bottomrule
\end{tabular}
\end{minipage}
\hfill
\begin{minipage}[t]{0.58\textwidth}
\centering
\textbf{(b) chrF precision/recall decomposition}\\[3pt]
\begin{tabular}{llrrrr}
\toprule
\textbf{Task} & \textbf{System} & \textbf{Tokens} & \textbf{chrP} & \textbf{chrR} & \textbf{chrF} \\
\midrule
JA-P$\rightarrow$EN-P & Zero-shot & 2015.1 & 48.45 & 66.24 & 61.71 \\
 & QLoRA & 1781.4 & 52.17 & 62.70 & 60.26 \\
JA-G$\rightarrow$EN-G & Zero-shot & 1245.7 & 68.42 & 56.61 & 58.63 \\
 & QLoRA & 1202.4 & 70.58 & 55.87 & 58.30 \\
JA-P$\rightarrow$EN-G & Zero-shot & 1847.7 & 46.56 & 61.25 & 57.61 \\
 & QLoRA & 1449.0 & 52.91 & 55.37 & 54.86 \\
\bottomrule
\end{tabular}
\end{minipage}
\caption{Additional diagnostics. Panel (a) changes only the requested target stage in a separately generated paired Qwen3.6-27B rerun on the same 1,000 \texttt{en\_pre$\rightarrow$ja\_granted} examples; the cross-stage row is its internal paired baseline and is distinct from the main benchmark run. Panel (b) compares Qwen3.6-27B zero-shot and QLoRA on the English-target tasks; Tokens is mean generated length, and chrP/chrR are character $n$-gram precision and recall.}
\label{tab:diagnostics}
\end{table*}

\begingroup
\raggedbottom
\section{Prompts and Implementation Details}
\label{app:prompts}
The verbatim prompt templates and implementation settings for all conditions are listed below; line breaks within the display blocks are typographical.

\subsection{Zero-shot}
\promptheading{System message}
\begin{PromptBox}
You are a professional patent translator
and patent claim drafter. Output only
the requested target claims.
\end{PromptBox}

\promptheading{Task-specific instructions}
\taskprompt{ja\_pre$\rightarrow$en\_pre}{Translate the following Japanese pre-grant patent application claims into English pre-grant patent application claims. Preserve the technical meaning, claim numbering, and legal scope.}
\taskprompt{ja\_pre$\rightarrow$en\_granted}{Translate and draft the following Japanese pre-grant patent application claims into English granted patent claim style. Preserve all technically and legally supported limitations, while allowing target-side claim consolidation or dependency reorganization when appropriate.}
\taskprompt{ja\_granted$\rightarrow$en\_granted}{Translate the following Japanese granted patent claims into English granted patent claim style. Preserve the final legal scope and use formal English patent drafting language.}
\taskprompt{en\_pre$\rightarrow$ja\_pre}{Translate the following English pre-grant patent application claims into Japanese pre-grant patent application claims. Preserve the application-claim breadth and technical meaning.}
\taskprompt{en\_pre$\rightarrow$ja\_granted}{Translate and draft the following English pre-grant patent application claims into Japanese granted patent claim style. Preserve all technically and legally supported limitations, while allowing target-side claim consolidation or dependency reorganization when appropriate.}

\promptheading{Output-format rule (selected by target language)}
\begin{PromptBox}
[Japanese target]
Output only the Japanese target patent
claims. Do not include explanations,
markdown, headings, or notes. Start
directly with claim 1 using numbered
claim format.

[English target]
Output only the English target patent
claims. Do not include explanations,
markdown, headings, or notes. Start
directly with claim 1 using the
format '1.'.
\end{PromptBox}

\promptheading{User-message template}
\begin{PromptBox}
{instruction}

{output_rule}

Now translate this input.

SOURCE CLAIMS:
{source_text}

TARGET CLAIMS:
\end{PromptBox}

\subsection{Few-shot}
\promptheading{System message}
\begin{PromptBox}
You are a professional patent translator
and patent claim drafter. Use the
examples only to learn category-specific
drafting style. Do not copy facts from
the examples. Output only the requested
target claims.
\end{PromptBox}

\promptheading{User-message template (one example per task)}
\begin{PromptBox}
{instruction}

{output_rule}

Category-specific examples follow. Use
them for drafting style only; do not
copy their facts.

Example 1
SOURCE CLAIMS:
{example_source}

TARGET CLAIMS:
{example_target}

Now translate this input.

SOURCE CLAIMS:
{source_text}

TARGET CLAIMS:
\end{PromptBox}
Each few-shot condition uses one fixed demonstration per task direction, reused across all evaluation instances; the same five demonstrations are used for the Qwen3.6-27B and GPT-5.4 few-shot conditions. Demonstrations were selected manually to be typical of their direction: for same-stage directions, a pair whose source and target claim sets have comparable token length and claim count; for cross-stage directions, a pair showing a moderate reduction in claim count from source to target (for example, thirty source claims to twenty target claims), rather than atypical cases with much larger expansions or reductions. The demonstrations are included in the released dataset.

\subsection{Simple-MT ablation}
\promptheading{System message}
\begin{PromptBox}
You are a professional technical
translator. Output only the requested
translation.
\end{PromptBox}
The instruction for all Japanese-source categories is ``Translate the following Japanese patent claims into English. Preserve the claim numbering and technical meaning.''; English-source categories use the mirrored Japanese-target wording.
The user-message assembly is identical to the zero-shot template.

\subsection{QLoRA and GPT-5.4 conditions}
The QLoRA condition and the GPT-5.4 API baseline use the identical zero-shot prompt template; only the model weights (via a LoRA adapter) or the serving API differ.

\subsection{Inference settings}
All system outputs are scored with the same implementation: sacreBLEU~\cite{post2018sacrebleu} for BLEU and chrF, with MeCab tokenization for Japanese-target BLEU, and the \texttt{evaluate} implementation for SARI. Reported values are rounded to two decimal places. BLEU and chrF are computed as corpus-level metrics over all instances, whereas SARI is computed separately for each instance and macro-averaged across instances.

Google Translate uses Google Cloud Translation API (Advanced) v3, accessed in June 2026. Inputs exceeding the API limit are divided into chunks and reassembled after translation.

Qwen3.6-27B inference uses vLLM~\cite{kwon2023vllm} with a maximum model length of 32,768 tokens, a maximum of 4,096 newly generated tokens, and greedy decoding with temperature 0. The zero-shot, few-shot, simple-MT, prompt-control, and QLoRA conditions use the same generation limit and decoding configuration.

The GPT-5.4 API conditions were accessed in July 2026 with reasoning disabled (\texttt{reasoning.effort=none}), temperature 0, and a 4,096-token output limit. The zero-shot and few-shot conditions use the same prompt templates and task-specific demonstration examples as the corresponding Qwen conditions. Both stages of the GPT-5.4 pipeline baselines use these settings.

\subsection{QLoRA training details}
Training uses an 8,192-token maximum sequence length and 4-bit NF4 quantization with double quantization and bfloat16 computation. LoRA adapters~\cite{hu2022lora} with rank 16, scaling factor 32, and dropout 0.05 are attached to the query, key, value, and output projection modules.

The per-device batch size is one with eight gradient-accumulation steps. Training uses \texttt{paged\_adamw\_8bit} for 12,000 steps with a peak learning rate of $2\times10^{-4}$, cosine decay, and a warmup ratio of 0.03. Validation and checkpointing occur every 3,000 steps. The final 12,000-step adapter is frozen during final evaluation.

\subsection{Pipeline prompts}
The pipeline drafting stages use a monolingual pre-grant-to-granted rewriting prompt, adapted from the claim-rewriting framing of \citet{kawano2024claimbrush} with its refusal-reason input omitted, since our dataset carries no Office Action metadata; the translation stages reuse the simple-MT prompt above verbatim.

\promptheading{System message}
\begin{PromptBox}
You are a professional patent claim
drafter. Output only the requested
target claims.
\end{PromptBox}

\promptheading{Drafting instruction (English variant)}
The Japanese variant is identical with ``English'' replaced by ``Japanese'' throughout.
\begin{PromptBox}
Draft the following English pre-grant
patent application claims into English
granted patent claim style, the way an
applicant amends claims during patent
examination so they are more likely to
be granted, rather than a general
paraphrase. Preserve all technically
and legally supported limitations and
existing claim numbering, while
allowing claim consolidation or
dependency reorganization when
appropriate. Do not translate into
another language; keep the output in
English.
\end{PromptBox}

The output rules are those of the zero-shot template, selected by the stage's output language.
The user-message assembly is identical to the zero-shot template with ``Now translate this input.'' replaced by ``Now rewrite this input.''.
\endgroup

\section{Procedure of Dataset Construction}
\label{app:appendix_construction}
Instances were selected at random without replacement. Each direction's evaluation sample spans 961, 964, 972, and 924 distinct patent families respectively, and 3,608 distinct families in total.

Figure~\ref{fig:construction} shows the overall process; this appendix records the concrete parameters.
We first combine the cleaned Japanese and English records, normalize their pre-grant and granted publication dates, determine when each document became publicly available, and remove duplicate records with the same stage pairing and Japanese/English claim text.
Family identifiers use the \texttt{family\_id} field in Google Patents Public Data (\path{patents-public-data.patents.publications}), provided by IFI CLAIMS Patent Services, and are resolved by joining each Japanese document's PCT application number, with PCT publication number as fallback, against that table~\cite{googlepatentsdata}.
For each PCT identifier, the Japanese side retains the claim-set document with the largest extracted claim count, breaking ties by ascending publication number, whereas the English side retains every qualifying USPTO publication in the family rather than one representative per family.
The source and target text fields contain complete claim sets without abstracts or descriptions.

% Patent families are then ordered chronologically and assigned to train, validation, and test partitions in an 80/10/10 ratio at the family level, preventing related records from crossing partitions.

Patent families are ordered by a family anchor date, defined as the latest of the available canonical pre-grant-visibility and grant dates across the family's Japanese and English records, with ties broken by ascending family identifier, and are assigned to train, validation, and test partitions in an 80/10/10 ratio at the family level.

We construct the six task directions in Table~\ref{tab:dataset_stats}; for cross-stage tasks, an instance is retained only when the source document was publicly available before the target grant.
Combined input--target lengths are measured with the Gemma tokenizer (\texttt{google/gemma-4-26B-A4B-it}), and instances exceeding 24k tokens are excluded, leaving 4,067,168 directional instances.
Each released instance carries publication numbers, publication dates, application numbers, and patent-family identifiers as metadata.
Finally, we sample test examples per task direction for the four core benchmark tasks and the supplementary granted-to-granted analysis.

\section{Original Japanese Claims for the Main-Text Comparison}
\label{app:claim_gap_japanese}

Table~\ref{tab:claim_gap_japanese} reproduces the original JPO claim text corresponding to the English translations in Table~\ref{tab:claim_gap}.

\begin{table}[h!]
\centering
\scriptsize
\setlength{\tabcolsep}{3pt}
\renewcommand{\arraystretch}{1.0}
\begin{tabular}{l|p{0.73\columnwidth}}
\hline
\textbf{Stage} & \textbf{Original Japanese claim} \\
\hline\hline
\textbf{Pre-grant} &
\texttt{JP-2004501492-A} (2004-01-15), Claim~1
\newline
\begin{JPtext}電解質層(2)と、前記電解質層を挟む一対のガス拡散電極層(3,4)と、前記各ガス拡散電極層に接する燃料ガス及び酸化剤ガスの通路(10,11)を画定するための一対の配流板(5)とを有する少なくとも1つのセルから構成された燃料電池であって、前記電解質層(2)が、フレーム(21)及び該フレーム内に保持された電解質(22)を有し、前記両配流板と前記フレームとが、内部が気密になるようにその外周全周に亘り接合または接着されていることを特徴とする燃料電池。\end{JPtext}
\\
\hline
\textbf{Granted} &
\texttt{JP-4989802-B2} (2012-08-01), Claim~1
\newline
\begin{JPtext}電解質層(2)と、前記電解質層を挟む一対のガス拡散電極層(3,4)と、前記各ガス拡散電極層に接する燃料ガス及び酸化剤ガスの通路(10,11)を画定するための一対の配流板(5)とを有する少なくとも1つのセルから構成された燃料電池であって、前記電解質層(2)が、フレーム(21)及び該フレーム内に保持された電解質(22)を有し、前記両配流板と前記フレームとが、内部が気密になるようにその外周全周に亘り、\jpbold{前記両配流板の外周部に成膜された電極膜と前記フレームの外周部に成膜されたガラス膜とにより陽極接合され、前記電極膜の下層に陽極接合のためのヒータが埋設されている}ことを特徴とする燃料電池。\end{JPtext}
\\
\hline
\end{tabular}
\caption{Original Japanese JPO claims for the translations in Table~\ref{tab:claim_gap}; bold marks the limitation added in the granted claim.}
\label{tab:claim_gap_japanese}
\end{table}
\FloatBarrier

\section{Pilot Pairwise Evaluation}
\label{app:human_eval}

The evaluation covers the \texttt{en\_pre$\rightarrow$ja\_granted} direction. Starting from the 1,000 benchmark evaluation rows, we required a usable auxiliary Japanese pre-grant claim set, excluded the five families used as few-shot demonstrations, removed artifact-containing or degenerate texts, required non-degenerate outputs from all three evaluated systems, and restricted the English source and Japanese granted reference to at most 2,500 Qwen3.6 tokens. A total of 449 rows survived these filters.

We sampled 50 cases using candidate-selection and divided them into two groups of 25. The reported evaluation uses Group A, in which an auxiliary Japanese pre-grant claim set was shown for terminology and technical understanding. Resemblance to this auxiliary text was explicitly not an evaluation criterion. Auxiliary-text repairs and deterministic replacements are documented in the released sampling manifest.

Each case produces three comparisons: Qwen simple-MT versus Qwen QLoRA, GPT-5.4 zero-shot versus Qwen QLoRA, and Qwen simple-MT versus GPT-5.4 zero-shot. Pair sides were randomized using seed 44, producing 75 judgments. System identities were hidden from the evaluator, and ties were permitted when there was no practically meaningful difference.

Qwen simple-MT is a translation-only prompting ablation using the same Qwen3.6-27B base model as QLoRA, whereas GPT-5.4 zero-shot provides a separate general-purpose API baseline. Each item presents the complete English pre-grant source claim set, the auxiliary Japanese pre-grant claim set, and two anonymized candidate outputs.

%The human evaluator was instructed to select the candidate expected to require fewer revisions as a Japanese granted-stage claim set. The instruction emphasized the following criteria:
The human evaluator was instructed to judge which candidate could be used in practice as a Japanese granted-stage claim set, as-is or with less revision. The instruction emphasized the following criteria:
\begin{quote}
%\criteriabox{
\small
%Prioritize technical errors, loss of invention identity, and unsupported
%additions; next consider clarity and claim-structure defects; do not
%decide based only on minor stylistic differences. Supported
%incorporation of limitations, claim consolidation, and dependency
%reorganization are not inherently errors. Do not assess novelty,
%inventive step, or actual grantability.
Prioritize technical errors, loss of invention identity, and clearly
inappropriate additions; next consider clarity and claim-structure
defects; do not decide based only on minor stylistic differences.
Supported incorporation of limitations, claim consolidation, and
dependency reorganization are not inherently errors. Do not assess
novelty, inventive step, actual grantability, or compliance with
amendment requirements.
%}
\end{quote}

The LLM evaluator (Fable 5) received the same items, instructions, and rubric as the human evaluator, evaluated each pair independently using the provider's default decoding parameters, and returned a structured verdict (first candidate, second candidate, or tie) with a brief rationale. The judge model belongs to a different model family from all evaluated systems. Its verdicts were held server-side and never displayed before the corresponding human decision was committed, so neither evaluator had access to the other's verdict when making a decision. We report human--LLM agreement as a diagnostic and do not treat the LLM evaluator as a substitute for human evaluation.

All evaluation materials (instructions, sampling manifest, items, raw judgments, and analysis) are available in the released repository.\footnote{\url{https://doi.org/10.6084/m9.figshare.33027056.v1}}

\newpage
\clearpage
\onecolumn
\section{Qualitative Output Comparison}
\label{app:qualitative}

\begin{table}[h!]
\centering
\footnotesize
\setlength{\tabcolsep}{5pt}
\renewcommand{\arraystretch}{0.97}
\begin{tabular}{@{}p{0.15\textwidth}p{0.80\textwidth}@{}}
\toprule

\multicolumn{2}{@{}l}{\textbf{(a) Successful consolidation: EN-P$\rightarrow$JA-G}} \\
\midrule

\textbf{Source}
\newline
20 claims
\newline
Claim 8
&
The producing method \textbf{according to claim 1}, wherein the protective film shows a main peak of a ratio (tan $\delta$) between a storage modulus and a loss modulus in dynamic viscoelasticity measurement at a temperature of 80$^\circ$C or higher.
\\
\addlinespace[3pt]

\textbf{Reference}
\newline
14 claims
\newline
Claim 8
&
\begin{JPtext}
前記保護フィルムは、動的粘弾性測定における貯蔵弾性率と損失弾性率との比(tan$\delta$)の主ピークを示す温度が80°C以上であることを含む、\jpbold{請求項1〜請求項7のいずれか1項}に記載の製造方法。
\end{JPtext}
\\
\addlinespace[3pt]

\textbf{Google Translate}
\newline
20 claims
\newline
chrF 53.31
\newline
Claim 8
&
\begin{JPtext}
\jpbold{請求項1に記載の方法}であって、前記保護フィルムは、80℃以上の温度における動的粘弾性測定において、貯蔵弾性率と損失弾性率との比（tan $\delta$）の主ピークを示すことを特徴とする製造方法。
\end{JPtext}
\\
\addlinespace[3pt]

\textbf{Qwen QLoRA}
\newline
14 claims
\newline
chrF 60.31
\newline
Claim 8
&
\begin{JPtext}
前記保護フィルムが、動的粘弾性測定において80°C以上の温度で貯蔵弾性率と損失弾性率との比(tan$\delta$)の主ピークを示すことを特徴とする\jpbold{請求項1〜7のいずれか一項}に記載の製造方法。
\end{JPtext}
\\

\midrule
\multicolumn{2}{@{}l}{\textbf{(b) Source-supported claim narrowing: JA-P$\rightarrow$EN-G}} \\
\midrule

\textbf{Source}
\newline
10 claims
\newline
Claims 1--2
&
\begin{JPtext}
1. セルロースエーテル及び尿素粉末を含むセメントモルタル用添加剤であって、前記添加剤に疎水性シリカが含まれていることを特徴とする、尿素粉末を含むセメントモルタル用添加剤。\newline
2. セルロースエーテルは\jpbold{65重量\%〜89.3重量\%}、尿素粉末は\jpbold{10重量\%〜30重量\%}、疎水性シリカは\jpbold{0.7重量\%〜5重量\%}の範囲であることを特徴とする、請求項1に記載の尿素粉末を含むセメントモルタル用添加剤。
\end{JPtext}
\\
\addlinespace[3pt]

\textbf{Reference}
\newline
11 claims
\newline
Claim 1
&
A cement mortar additive comprising cellulose ether, urea powder, and hydrophobic silica, wherein the content of the cellulose ether is \textbf{65--89.3 wt\%}, the content of the urea powder is \textbf{10--30 wt\%}, and the content of the hydrophobic silica is \textbf{0.7--5 wt\%}.
\\
\addlinespace[3pt]

\textbf{Google Translate}
\newline
10 claims
\newline
chrF 67.00
\newline
Claims 1--2
&
1. A cement mortar additive containing cellulose ether and urea powder, characterized in that the additive contains hydrophobic silica.\newline
2. The cement mortar additive according to Claim~1, wherein the cellulose ether is \textbf{65--89.3 wt\%}, the urea powder is \textbf{10--30 wt\%}, and the hydrophobic silica is \textbf{0.7--5 wt\%}.
\\
\addlinespace[3pt]

\textbf{Qwen QLoRA}
\newline
10 claims
\newline
chrF 78.57
\newline
Claim 1
&
An additive for cement mortar comprising urea powder, comprising cellulose ether, urea powder, and hydrophobic silica, wherein the cellulose ether is present in an amount of \textbf{65--89.3 wt\%}, the urea powder is present in an amount of \textbf{10--30 wt\%}, and the hydrophobic silica is present in an amount of \textbf{0.7--5 wt\%}.
\\

\midrule
\multicolumn{2}{@{}l}{\textbf{(c) Over-consolidation failure: JA-P$\rightarrow$EN-G}} \\
\midrule

\textbf{Source}
\newline
18 claims
\newline
4 branches
\newline
Claim 6
&
\begin{JPtext}
その必要がある動物の\jpbold{脂肪組織における脂肪分解を増大させる方法}であって、約0.5gから約30gまでの$\beta$−ヒドロキシ−$\beta$−メチル酪酸(HMB)の組成物を動物に投与し、それによって動物の脂肪分解を増大させることを含む方法。
\end{JPtext}
\\
\addlinespace[3pt]

\textbf{Reference}
\newline
16 claims
\newline
4 branches
\newline
Claim 1
&
A method of \textbf{increasing intracellular lipolysis in adipose tissue} of an animal in need thereof comprising administering to said animal a composition of from about 0.5 g to about 30 g of $\beta$-hydroxy-$\beta$-methylbutyric acid (HMB), thereby increasing lipolysis in the adipose tissue of the animal.
\\
\addlinespace[3pt]

\textbf{Google Translate}
\newline
18 emitted claims
\newline
4 branches
\newline
chrF 70.71
\newline
Claim 6
&
A method for \textbf{increasing lipolysis in the adipose tissue} of an animal in need, comprising administering to the animal a composition of $\beta$-hydroxy-$\beta$-methylbutyrate (HMB) in an amount from about 0.5 g to about 30 g, thereby increasing lipolysis in the animal.
\\
\addlinespace[3pt]

\textbf{Qwen QLoRA}
\newline
5 claims
\newline
1 branch
\newline
chrF 26.32
\newline
Claim 1
&
A method of \textbf{reducing body fat} in an animal comprising administering to the animal a composition comprising about 0.5 g to about 30 g of $\beta$-hydroxy-$\beta$-methylbutyric acid (HMB), wherein administration of the composition increases lipolysis in the body of the animal.
\\

\bottomrule
\end{tabular}
\caption{Qualitative comparison of Google Translate and QLoRA against the reference. Panel~(a) shows reference-like consolidation; panel~(b), source-supported narrowing by incorporating source Claim~2's ranges into independent Claim~1; and panel~(c), over-consolidation in which QLoRA reduces four method branches to one and omits the glycerol-export and comorbidity branches. Examples are illustrative, not assessments of legal validity; claim numbering differs because outputs are complete claim sets. P and G denote pre-grant and granted stages.}

\label{tab:qualitative}
\end{table}

\end{document}